\documentclass[conference,a4paper]{IEEEtran}
\IEEEoverridecommandlockouts
\usepackage{cite}
\usepackage{amsmath,amssymb,amsfonts}
\usepackage{algorithmic}
\usepackage{graphicx}
\usepackage{textcomp}
\usepackage{xcolor}
\usepackage{caption}
\usepackage{subcaption}
\usepackage{booktabs}
\usepackage{url}
\usepackage{hyperref}

\usepackage{makecell}
\usepackage{enumitem}
\setlist[itemize]{noitemsep, topsep=0pt, parsep=0pt, partopsep=0pt}
\usepackage{setspace}

\usepackage{fancyhdr}
\fancypagestyle{firstpage}{
  \fancyhf{}                         
  
  \fancyfoot[C]{%
    \footnotesize
    \parbox{\textwidth}{\centering
      \textcopyright~2026 IEEE. Personal use of this material is permitted.
      Permission from IEEE must be obtained for all other uses, in any current or future media, including reprinting/republishing this material for advertising or promotional purposes, creating new collective works, for resale or redistribution to servers or lists, or reuse of any copyrighted component of this work in other works.\\[0.3em]
    }%
  }
}
\def\BibTeX{{\rm B\kern-.05em{\sc i\kern-.025em b}\kern-.08em
    T\kern-.1667em\lower.7ex\hbox{E}\kern-.125emX}}
\begin{document}
\bstctlcite{IEEEexample:BSTcontrol}
\title{NeRV360: Neural Representation for 360-Degree Videos with a Viewport Decoder
}

\author{\IEEEauthorblockN{Daichi Arai, Kyohei Unno, Yasuko Sugito, and Yuichi Kusakabe}
  \IEEEauthorblockA{\textit{Science \& Technology Research Laboratories}, \textit{NHK}, Tokyo, Japan\\
    \{arai.d-es, unno.k-iw, sugitou.y-gy, kusakabe.y-ee\}@nhk.or.jp}}

\maketitle

\thispagestyle{empty} 

\thispagestyle{firstpage}

\begin{abstract}
Implicit neural representations for videos (NeRV) have shown strong potential for video compression. However, applying NeRV to high-resolution 360-degree videos causes high memory usage and slow decoding, making real-time applications impractical. We propose \textbf{NeRV360}, an end-to-end framework that decodes only the user-selected viewport instead of reconstructing the entire panoramic frame. Unlike conventional pipelines, NeRV360 integrates viewport extraction into decoding and introduces a spatio-temporal-aware affine transform module for conditional decoding based on viewpoint and time. Experiments on 6K-resolution videos show that NeRV360 achieves a \(7 \times\) reduction in memory consumption and a \(2.5 \times\) increase in decoding speed compared to HNeRV, a representative prior work, while delivering better image quality in terms of objective metrics.
\end{abstract}


\section{Introduction}

360-degree video content has become increasingly popular in applications for virtual reality \cite{matoba2019vr8k, eltobgy2020mobile}. Viewing devices, such as head-mounted displays or touchscreen displays, typically show only a limited viewport at any given time, unlike conventional videos. To maintain visual quality in these viewports, high-resolution 360-degree video is essential because viewports rendered from the entire panoramic frame contain far fewer pixels. However, increasing resolution also significantly increases video data size, necessitating more efficient compression techniques.
 
To address the growing demand for efficient video compression, end-to-end video compression approaches \cite{jia2025towards, zhang2025flavc, regensky2025beyond, arai2025neural, guo2025generative} have emerged. Among these, implicit neural representations for videos (NeRV) \cite{NEURIPS2021_b4418237, li2022nerv, 10205438, 10658449, wu2024qs,wei2024snerv, 10871929, 10566345, gao2025pnvc, chen2024fast, 10849907, kwan2024nvrc, gao2025givic, fujihashi2024fv, hayami2025neural, ling2025multi} have delivered competitive compression performance. NeRV-based methods encode a target video by overfitting a neural network and subsequently decode it through a feedforward inference process, offering a simple yet effective alternative to conventional codecs. The original NeRV framework \cite{NEURIPS2021_b4418237}, which uses frame indices as input, has undergone substantial development, evolving into HNeRV \cite{10205438}, where a ConvNeXt-based encoder \cite{liu2022convnet} generates embeddings as input. Further improvements were introduced by Boosting-NeRV \cite{10658449}, which employs a conditional decoder with a temporal-aware affine transform (TAT) module. The current state-of-the-art, GIViC \cite{gao2025givic}, achieves better compression efficiency than Versatile Video Coding (VVC) \cite{9503377}, demonstrating the strong potential of NeRV-based approaches.

\begin{figure}
    \centering
    \includegraphics[width=1\linewidth]{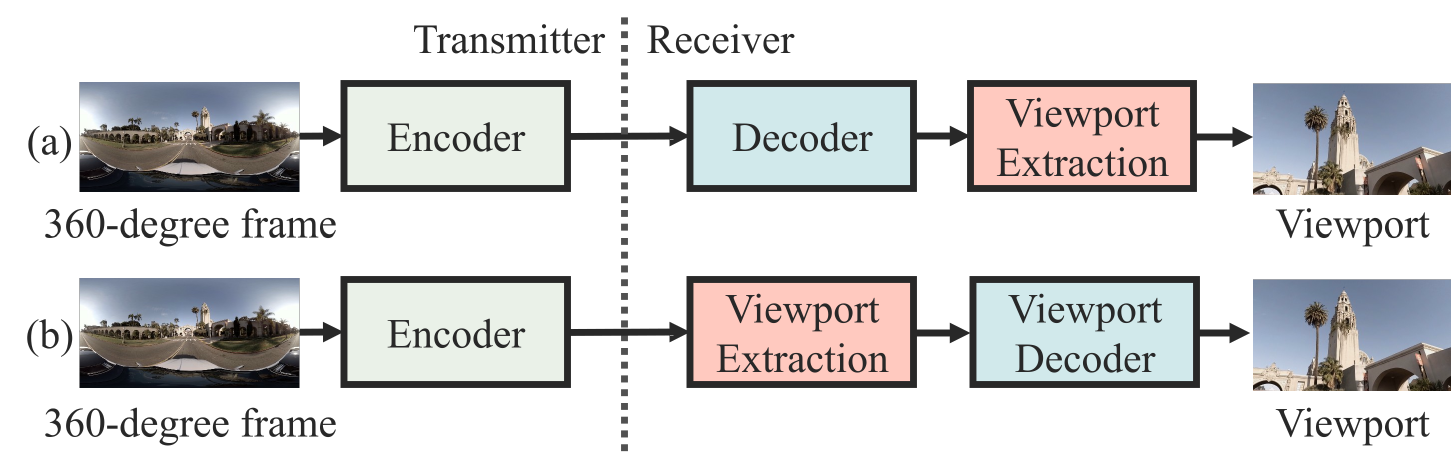}
    \vspace{-15pt}
    \caption{Comparison of pipelines: (a) conventional decoding followed by viewport extraction \cite{matoba2019vr8k}, and (b) NeRV360 decoding with integrated viewport extraction.}
    \label{fig:NeRV360_traditional_vs_proposed}
    \vspace{-15pt}
\end{figure}

Despite these advances, applying NeRV to high-resolution videos remains challenging due to substantial memory requirements and limited decoding speed. Prior works have mainly targeted resolutions up to 2K, often overlooking scalability issues for ultra-high-resolution content such as 360-degree videos. Prior approaches decode the entire panoramic frame before extracting the viewport, resulting in high computational cost and making real-time processing infeasible on GPUs with limited memory. For example, even with half-precision computation and a compact model size of 2.2M for 6K-resolution sequences, HNeRV-Boost \cite{10658449} requires approximately 30\,GiB of GPU memory for decoding, making real-time applications impractical.

\begin{figure*}
  \begin{subfigure}{\columnwidth}
    \centering
    \includegraphics[width=2\columnwidth]{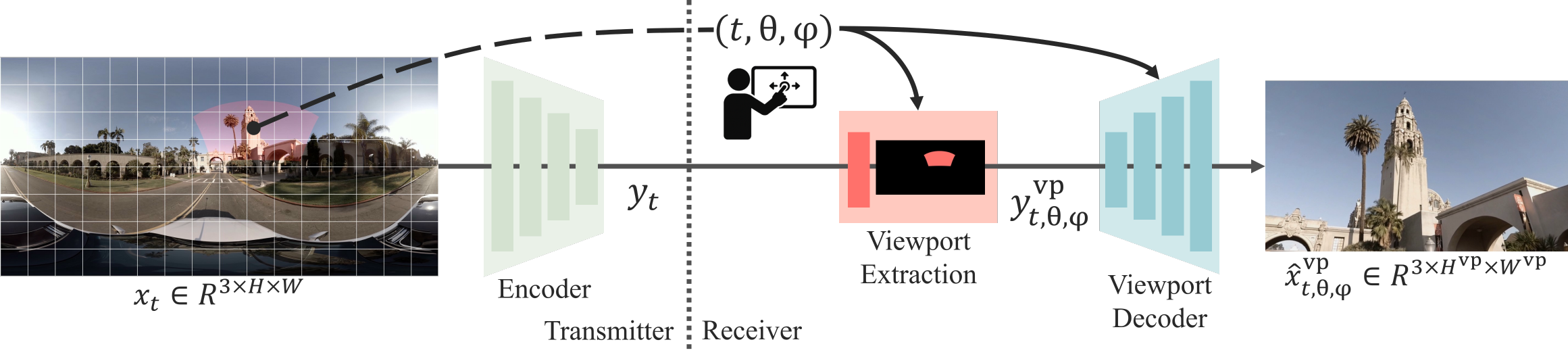}
    \vspace{-15pt}
    \captionsetup{labelformat=empty}
    \label{fig:1a}
  \end{subfigure}
  \\
  \begin{subfigure}{\columnwidth}
    \centering
    \includegraphics[width=2\columnwidth]{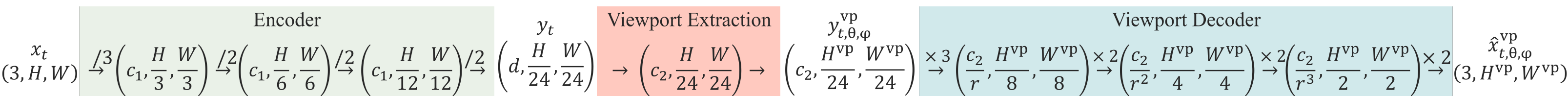}
    \vspace{-15pt}
    \captionsetup{labelformat=empty}
    \label{fig:1b}
  \end{subfigure}
    \caption{Overview of the NeRV360 framework. The input 360-degree frame $x_t$ is represented as $(3, H, W)$ and the output viewport $\hat{x}^{\mathrm{vp}}_{t,\theta,\varphi}$ as $(3, H^{\mathrm{vp}}, W^{\mathrm{vp}})$. The encoder and decoder strides are $(3, 2, 2, 2)$.}
  \vspace{-15pt}
  \label{fig:overview}
\end{figure*}

To address this challenge, we propose \textbf{NeRV360}, an end-to-end neural representation specifically designed for 360-degree videos. As shown in Fig.~\ref{fig:NeRV360_traditional_vs_proposed}, our approach transmits the entire 360-degree video and extracts viewports on the receiver side, unlike viewport-adaptive and tile-based streaming methods that transmit only part of the content on the transmitter side. Conventional video compression methods typically decode the entire panoramic frame before extracting the viewport, introducing unnecessary overhead because viewing devices such as head-mounted displays and touchscreen displays show only a limited portion of the panoramic view. Inspired by the joint rescaling and viewport rendering approach \cite{li2024resvr}, NeRV360 eliminates this inefficiency by integrating viewport extraction directly into the decoding process, enabling selective reconstruction of only the user-selected viewport within the embedding space. This design significantly reduces computational cost on the decoder side, allowing faster decoding and efficient training of high-resolution 360-degree video models on commonly available GPUs.

This study makes the following key contributions to viewport decoding:
\begin{itemize}
\item We introduce a viewport decoder that reconstructs viewports directly without decoding the entire frame.
\item We incorporate a channel expansion layer before viewport extraction to mitigate quality degradation caused by bilinear interpolation in the embedding space.
\item We introduce a viewpoint-conditioned mechanism using longitude, latitude, and temporal embeddings.
\end{itemize}


\section{Methodology}

\subsection{Overview}

As illustrated in Fig.~\ref{fig:overview}, our 360-degree video compression pipeline takes the input frame \( x_t \) in equirectangular format, where \( t \) denotes the frame index. The output viewport is represented as \( \hat{x}^{\mathrm{vp}}_{t,\theta,\varphi} \), with \( \theta \) and \( \varphi \) indicating the longitude and latitude of the viewport center. These parameters are user-defined, allowing viewers to freely select any direction within the 360-degree scene.
Conventional encoder–decoder models such as HNeRV \cite{10205438} first process the input frame \( x_t \) to generate the embedding \( y_t \), which is then decoded into the entire panoramic frame \( \hat{x}_t \) before extracting the user-selected viewport \( \hat{x}^{\mathrm{vp}}_{t,\theta,\varphi} \). In contrast, NeRV360 performs viewport extraction prior to decoding, achieving significant gains in memory efficiency and processing speed. Specifically, the viewpoint parameters \( \theta \) and \( \varphi \) are used to apply a perspective projection, extracting the corresponding viewport region \( y^{\mathrm{vp}}_{t,\theta,\varphi} \) directly from the embedding \( y_t \). To generate embeddings with sufficient spatial detail for this process, we adapt ConvNeXt \cite{liu2022convnet} as the encoder, similar to HNeRV and HNeRV-Boost \cite{10658449}.

\subsection{Viewport Extraction}

Similar to conventional viewport extraction techniques \cite{carroll2009optimizing, jabar2017perceptual}, our framework employs a perspective projection transformation, where the viewport is determined by the user-selected viewpoint, defined by longitude \( \theta \), latitude \( \varphi \), and the field of view, which is typically determined by the specifications of the viewing device. This transformation uses bilinear interpolation, which computes pixel values by interpolating the four nearest points on the equirectangular map corresponding to the viewport coordinates. However, bilinear interpolation in the embedding space can introduce blurriness due to weighted averaging, resulting in performance degradation. To address this limitation, we introduce a channel expansion layer that incorporates a sinusoidal NeRV-like (SNeRV) block and a TAT module for conditional decoding based on the frame index \( t \), both adapted from Boosting-NeRV \cite{10658449}. By increasing the channel dimension from \( d \) to \( c_2 \) prior to applying the perspective projection, our approach mitigates interpolation artifacts in the embedding space, thereby improving overall performance.

\begin{figure}
    \centering
    \includegraphics[width=1\linewidth]{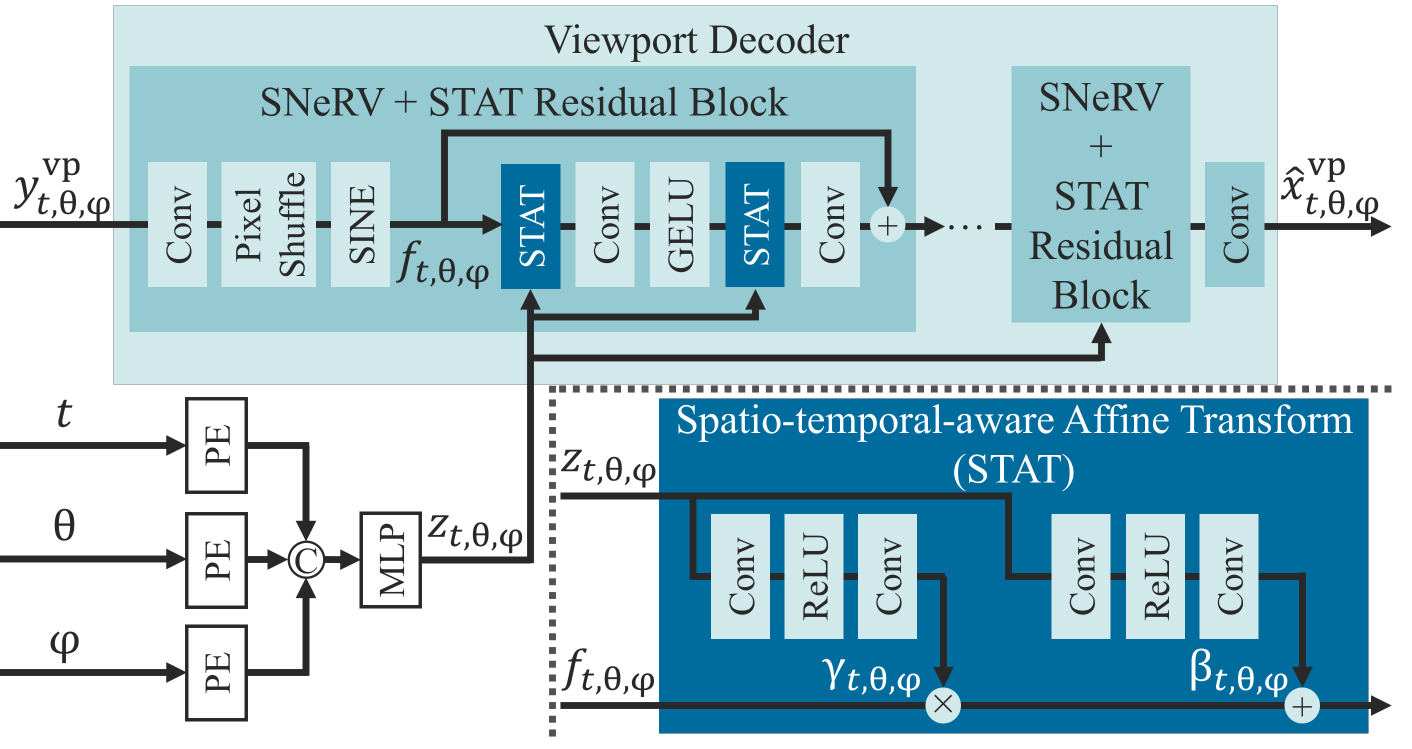}
    \caption{Illustration of our viewport decoder and STAT.}
    \label{fig:STAT}
    \vspace{-15pt}
\end{figure}

\begin{table*}[t]
\centering
\footnotesize
\renewcommand{\arraystretch}{1.2}
\caption{Comparison of regression performance on JVET Class S2 6K-resolution 360-degree videos.}
\begin{tabular}{lcccccccccc}
\toprule
\textbf{Model} & \textbf{Strides} & \textbf{Decoder size} & \multicolumn{2}{c}{\textbf{Dynamic viewport}} & \multicolumn{3}{c}{\textbf{Frame rate [FPS]}} & \multicolumn{3}{c}{\textbf{GPU memory [GiB]}} \\
\cmidrule(lr){4-5} \cmidrule(lr){6-8} \cmidrule(lr){9-11}
& & & PSNR & MS-SSIM & Encoding & Decoding & Training & Encoding & Decoding & Training \\
\midrule
HNeRV \cite{10205438} & $(3,2,2,2)$ & 2.2M & 24.37 & 0.728 & 27.3 & 15.0 & 2.7 & 2.8 & 26.2 & 51.3 \\
HNeRV-Boost \cite{10658449} & $(3,2,2,2)$ & 2.2M & 24.74 & 0.740 & 27.3 & 5.0 & 0.8 & 2.8 & 30.3 & 73.2 \\
NeRV360 & $(3,2,2,2)$ & 2.2M & 24.39 & 0.734 & 27.3 & 38.4 & 3.4 & 2.8 & 3.6 & 21.2 \\
\bottomrule
\end{tabular}
\label{tab:viewport_models_with_size}
\vspace{-10pt}
\end{table*}

\begin{table}[t]
\centering
\footnotesize
\renewcommand{\arraystretch}{1.2}
\caption{Ablation study results on the Balboa sequence.}
\begin{tabular}{llll}
\toprule
\textbf{Variant} & \multicolumn{1}{c}{\textbf{Embedding size}} & \multicolumn{2}{c}{\textbf{PSNR}} \\
\cmidrule(lr){3-4}
& \multicolumn{1}{c}{\textbf{($c \times h \times w$)}} & VP0 & VP1 \\
\midrule
NeRV360 & $1\times128\times256$ & \textbf{26.69} & \textbf{27.00} \\
\midrule
w/ 4$\times$ channels, $1/2$ resolution & $4\times64\times128$ & 26.34 & 26.66 \\
w/ 16$\times$ channels, $1/4$ resolution & $16\times32\times64$ & 24.74 & 24.89 \\
\midrule
w/ 1$\times$ channels, $1/2$ resolution & $1\times64\times128$ & 25.29 & 25.40 \\
w/ 1$\times$ channels, $1/4$ resolution & $1\times32\times64$ & 24.02 & 23.45 \\
\midrule
\makecell[l]{Channel expansion after \\ viewport extraction} & $1\times128\times256$ & 25.34 & 25.69 \\
w/o STAT inputs ($\theta, \varphi$) & $1\times128\times256$ & 26.51 & 26.85 \\
\bottomrule
\end{tabular}
\label{tab:merged_variant_embedding}
\vspace{-10pt}
\end{table}

\subsection{Viewport Decoder}

In Boosting-NeRV \cite{10658449}, temporal embeddings are derived from the frame index \( t \) using positional encoding (PE) and a multi-layer perceptron (MLP) for the TAT module. However, when viewport extraction is applied before decoding, the input embedding becomes viewpoint-dependent, adding complexity to the decoding process. To address this challenge, we propose a viewpoint-based spatio-temporal-aware affine transform (STAT) module, which extends TAT to incorporate longitude and latitude, as shown in Fig.~\ref{fig:STAT}. The STAT module learns affine parameters \(\beta_{t,\theta,\varphi}\) and \(\gamma_{t,\theta,\varphi}\) from time, latitude, and longitude embeddings, enabling conditional decoding. The feature transformation in the viewport decoder is expressed as:

\vspace{-10pt}
\begin{equation}
\text{STAT}(f_{t,\theta,\varphi} \mid \beta_{t,\theta,\varphi}, \gamma_{t,\theta,\varphi}) = \gamma_{t,\theta,\varphi} f_{t,\theta,\varphi} + \beta_{t,\theta,\varphi}.
\end{equation}
\vspace{-10pt}

\noindent The representation capability of the decoder is enhanced by alternately combining SNeRV blocks with STAT residual blocks that include Gaussian error linear units (GELUs) \cite{hendrycks2016gaussian}, improving robustness to variations in viewpoint location.


\section{Experiments}

\subsection{Experimental Setup}

We evaluated the effectiveness of our proposed framework using the JVET Class S2 test sequences \cite{he2021jvet}, which are employed for VVC exploration and consist of four videos, each with a frame size of \(3072 \times 6144\) pixels and durations of either 600 or 300 frames. Following the JVET common test conditions, we configured the output viewport size to \(1080 \times 1920\) pixels with a horizontal field of view of 78.1 degrees and employed dynamic viewport testing with two trajectories (VP0 and VP1) per sequence. For training on 6K-resolution 360-degree videos, we set the encoder and decoder strides to \((3, 2, 2, 2)\), resulting in an embedding size of \(128 \times 256\), which was further reduced to \(45 \times 80\) via perspective projection using randomly sampled \(\theta\) and \(\varphi\). The encoder channel \(c_1\) and embedding channel \(d\) were set to 64 and 1, respectively, and the channel reduction factor \(r\) was set to 1.2. In addition, we configured the positional encoding parameters to \(b = 1.25\) and \(l = 80\) for each of \(t\), \(\theta\), and \(\varphi\). The distortion loss function followed the same formulation and parameters as Boosting-NeRV \cite{10658449}:

\vspace{-10pt}
\begin{equation}
\begin{split}
L_d = & L_1(\text{FFT}(x^{\mathrm{vp}}_{t,\theta,\varphi}), \text{FFT}(\hat{x}^{\mathrm{vp}}_{t,\theta,\varphi})) + \lambda \alpha L_1(x^{\mathrm{vp}}_{t,\theta,\varphi}, \hat{x}^{\mathrm{vp}}_{t,\theta,\varphi}) + \\
& \lambda (1 - \alpha)(1 - L_{\text{MS-SSIM}}(x^{\mathrm{vp}}_{t,\theta,\varphi}, \hat{x}^{\mathrm{vp}}_{t,\theta,\varphi})),
\end{split}
\end{equation}
\vspace{-10pt}

\noindent where \(x^{\mathrm{vp}}_{t,\theta,\varphi}\) and \(\hat{x}^{\mathrm{vp}}_{t,\theta,\varphi}\) denote the original and decoded viewports, FFT represents the Fourier transform, and \(\lambda\) and \(\alpha\) were set to 60 and 0.7, respectively, with \(\lambda\) controlling the overall loss weight and \(\alpha\) balancing the contributions of L1 and MS-SSIM. We used Adan \cite{xie2024adan} as the optimizer with a cosine learning rate decay schedule and a warm-up phase comprising 10\% of the total epochs, starting from an initial learning rate of \(1.0 \times 10^{-4}\). For evaluation, we compared NeRV360 with HNeRV \cite{10205438} and HNeRV-Boost \cite{10658449}. Each model was trained for 300 epochs with a batch size of 1 using half-precision calculations. To ensure fairness, we set the strides to \((3, 2, 2, 2)\) so that all models share the same embedding size and standardized the model size to 2.2M. In addition, we measured the frame rate and GPU memory usage on an NVIDIA H100 GPU, where memory was obtained using the \texttt{torch.cuda.max\_memory\_allocated()} function in PyTorch to capture peak allocation.

\begin{figure}
    \centering
    \includegraphics[width=1\linewidth]{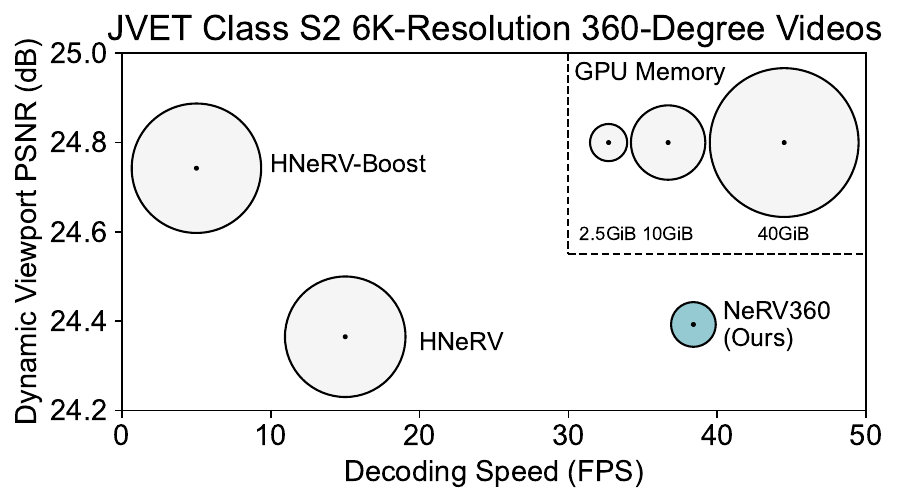}
    \vspace{-20pt}
    \caption{Comparison of NeRV360, HNeRV \cite{10205438}, and HNeRV-Boost \cite{10658449} for video regression with equivalent model sizes.
    }
    \vspace{-10pt}
    \label{fig:NeRV360_fps_vs_psnr}
\end{figure}

\subsection{Experimental Results}

Fig.~\ref{fig:NeRV360_fps_vs_psnr} and Table~\ref{tab:viewport_models_with_size} present regression results comparing HNeRV \cite{10205438}, HNeRV-Boost \cite{10658449}, and NeRV360 with the same decoder size on the JVET Class S2 dataset. The table summarizes average PSNR and MS-SSIM values, as well as frame rate and GPU memory usage during encoding, decoding, and training. Since the encoder was identical for all models, encoding frame rate and memory usage were the same. NeRV360 achieved a \(7 \times\) reduction in memory consumption and a \(2.5 \times\) increase in decoding speed compared to HNeRV, while delivering higher PSNR and MS-SSIM. Furthermore, training HNeRV and HNeRV-Boost on 6K-resolution sequences required over 50\,GiB of GPU memory even with a small decoder size of 2.2M, resulting in high costs for training high-quality models. In contrast, NeRV360 models of the same size could be trained on 6K-resolution sequences using consumer-grade GPUs with 24\,GiB of memory, making NeRV360 a cost-efficient solution for 360-degree video content delivery.

Table~\ref{tab:merged_variant_embedding} presents results from our ablation study using a decoder model of size 17.6M trained for 300 epochs on the Balboa sequence. Modifying encoder strides to reduce embedding resolution, even when maintaining overall size by increasing channel dimensions, results in inferior performance. Specifically, for equirectangular input frames at 6K-resolution and viewports at 2K-resolution, an embedding size of \(128 \times 256\) with a single channel (\(d = 1\)) yields better performance than configurations with higher channel dimensions. Furthermore, reducing resolution without fixing overall size causes a significant performance drop, indicating that embedding size is a crucial factor in our approach. Table~\ref{tab:merged_variant_embedding} also reports two additional ablation results: placing the channel expansion layer after viewport extraction and removing longitude and latitude inputs from STAT modules, confirming the effectiveness of both components.


\section{Conclusion}

This paper introduced NeRV360, an extension of the NeRV framework for 360-degree videos that enables direct viewport decoding. By integrating viewport extraction into the decoding process rather than performing it as a post-decoding step, our approach eliminates redundant decoding of non-visible regions. Experimental results demonstrate that NeRV360 achieves a \(7 \times\) reduction in memory consumption and a \(2.5 \times\) increase in decoding speed compared to HNeRV with the same decoder size.

Although our current framework supports pitch and yaw rotations, future work will explore extensions supporting roll and variable fields of view to enable more flexible viewport rendering. Its lightweight decoding design makes NeRV360 suitable for deployment on devices with limited computational capabilities. Furthermore, the ability to efficiently decode selected viewports enables promising applications for immersive experiences at resolutions exceeding 8K, where constraints on processing power and memory pose critical bottlenecks.


\bibliographystyle{IEEEtran}
\bibliography{refs}

@inproceedings{NEURIPS2021_b4418237,
 author = {Chen, Hao and He, Bo and Wang, Hanyu and Ren, Yixuan and Lim, Ser Nam and Shrivastava, Abhinav},
 booktitle = {{Advances in Neural Information Processing Systems}},
 pages = {21557--21568},
 title = {{NeRV: Neural Representations for Videos}},
 volume = {34},
 year = {2021}
}

@inproceedings{li2022nerv,
  title={{E-nerv: Expedite neural video representation with disentangled spatial-temporal context}},
  author={Li, Zizhang and Wang, Mengmeng and Pi, Huaijin and Xu, Kechun and Mei, Jianbiao and Liu, Yong},
  booktitle={{European Conference on Computer Vision}},
  pages={267--284},
  year={2022},
}

@INPROCEEDINGS{10205438,
  author={Chen, Hao and Gwilliam, Matthew and Lim, Ser-Nam and Shrivastava, Abhinav},
  booktitle={{Proceedings of the IEEE/CVF Conference on Computer Vision and Pattern Recognition (CVPR)}}, 
  title={{HNeRV: A Hybrid Neural Representation for Videos}}, 
  year={2023},
  volume={},
  number={},
  pages={10270-10279},
  doi={10.1109/CVPR52729.2023.00990}}

@INPROCEEDINGS{10658449,
  author={Zhang, Xinjie and Yang, Ren and He, Dailan and Ge, Xingtong and Xu, Tongda and Wang, Yan and Qin, Hongwei and Zhang, Jun},
  booktitle={{Proceedings of the IEEE/CVF Conference on Computer Vision and Pattern Recognition (CVPR)}}, 
  title={{Boosting Neural Representations for Videos with a Conditional Decoder}}, 
  year={2024},
  volume={},
  number={},
  pages={2556-2566},
  doi={10.1109/CVPR52733.2024.00247}}

@inproceedings{chen2024fast,
  title={{Fast encoding and decoding for implicit video representation}},
  author={Chen, Hao and Xie, Saining and Lim, Ser-Nam and Shrivastava, Abhinav},
  booktitle={{European Conference on Computer Vision}},
  pages={402--418},
  year={2024},
}

@inproceedings{gao2025pnvc,
  title={{Pnvc: Towards practical inr-based video compression}},
  author={Gao, Ge and Kwan, Ho Man and Zhang, Fan and Bull, David},
  booktitle={{Proceedings of the AAAI Conference on Artificial Intelligence}},
  volume={39},
  number={3},
  pages={3068--3076},
  year={2025}
}

@INPROCEEDINGS{10849907,
  author={Li, Hao and Yu, Lu and Liao, Yiyi},
  booktitle={Proceedings of the IEEE International Conference on Visual Communications and Image Processing (VCIP)}, 
  title={{PET-NeRV: Bridging Generalized Video Codec and Content-Specific Neural Representation}}, 
  year={2024},
  volume={},
  number={},
  pages={1-5},
  doi={10.1109/VCIP63160.2024.10849907}}

@article{kwan2024nvrc,
  title={{NVRC: Neural video representation compression}},
  author={Kwan, Ho Man and Gao, Ge and Zhang, Fan and Gower, Andrew and Bull, David},
  journal={Advances in Neural Information Processing Systems},
  volume={37},
  pages={132440--132462},
  year={2024}
}

@inproceedings{fujihashi2024fv,
  title={{FV-NeRV: Neural Compression for Free Viewpoint Videos}},
  author={Fujihashi, Takuya and Kato, Sorachi and Koike-Akino, Toshiaki},
  booktitle={Workshop on Machine Learning and Compression, NeurIPS},
  pages={1-10},
  year={2024}
}

@inproceedings{hayami2025neural,
  title={{Neural Video Representation for Redundancy Reduction and Consistency Preservation}},
  author={Hayami, Taiga and Shindo, Takahiro and Akamatsu, Shunsuke and Watanabe, Hiroshi},
  booktitle={{Proceedings of the IEEE International Conference on Consumer Electronics (ICCE)}},
  pages={1--6},
  year={2025},
}

@INPROCEEDINGS{10566345,
  author={Kwan, Ho Man and Zhang, Fan and Gower, Andrew and Bull, David},
  booktitle={{Proceedings of the Picture Coding Symposium (PCS)}}, 
  title={{Immersive Video Compression Using Implicit Neural Representations}}, 
  year={2024},
  volume={},
  number={},
  pages={1-5},
  doi={10.1109/PCS60826.2024.10566345}}

@ARTICLE{10871929,
  author={Zhu, Chen and Lu, Guo and He, Bing and Xie, Rong and Song, Li},
  journal={IEEE Transactions on Image Processing}, 
  title={{Implicit-Explicit Integrated Representations for Multi-View Video Compression}}, 
  year={2025},
  volume={34},
  number={},
  pages={1106-1118},
  doi={10.1109/TIP.2025.3536201}}

@inproceedings{wu2024qs,
  title={{QS-NeRV: Real-Time Quality-Scalable Decoding with Neural Representation for Videos}},
  author={Wu, Chang and Quan, Guancheng and He, Gang and Lai, Xin-Quan and Li, Yunsong and Yu, Wenxin and Lin, Xianmeng and Yang, Cheng},
  booktitle={{Proceedings of the 32nd ACM International Conference on Multimedia}},
  pages={2584--2592},
  year={2024}
}

@inproceedings{wei2024snerv,
  title={{SNeRV: Scalable Neural Representations for Video Coding}},
  author={Wei, Yiying and Amirpour, Hadi and Timmerer, Christian},
  booktitle={Workshop on Machine Learning and Compression, NeurIPS},
  year={2024}
}

@article{gao2025givic,
  title={{GIViC: Generative Implicit Video Compression}},
  author={Gao, Ge and Teng, Siyue and Peng, Tianhao and Zhang, Fan and Bull, David},
  journal={arXiv preprint arXiv:2503.19604},
  year={2025}
}

@inproceedings{jabar2017perceptual,
  title={{Perceptual analysis of perspective projection for viewport rendering in 360° images}},
  author={Jabar, Falah and Ascenso, Jo{\~a}o and Queluz, Maria Paula},
  booktitle={{Proceedings of the IEEE International Symposium on Multimedia (ISM)}},
  pages={53--60},
  year={2017},
}

@inproceedings{li2024resvr,
  title={{Resvr: Joint rescaling and viewport rendering of omnidirectional images}},
  author={Li, Weiqi and Zhao, Shijie and Chen, Bin and Cheng, Xinhua and Li, Junlin and Zhang, Li and Zhang, Jian},
  booktitle={Proceedings of the 32nd ACM International Conference on Multimedia},
  pages={78--87},
  year={2024}
}

@inproceedings{liu2022convnet,
  title={{A convnet for the 2020s}},
  author={Liu, Zhuang and Mao, Hanzi and Wu, Chao-Yuan and Feichtenhofer, Christoph and Darrell, Trevor and Xie, Saining},
  booktitle={{Proceedings of the IEEE/CVF conference on computer vision and pattern recognition}},
  pages={11976--11986},
  year={2022}
}

@article{hendrycks2016gaussian,
  title={{Gaussian error linear units (gelus)}},
  author={Hendrycks, Dan and Gimpel, Kevin},
  journal={arXiv preprint arXiv:1606.08415},
  year={2016}
}

@article{he2021jvet,
  title={{JVET Common Test Conditions and Evaluation Procedures for 360° Video}},
  author={He, Yuwen and Boyce, Jill and Choi, Kiho and Lin, Jian-Liang},
  journal={JVET-U2012},
  year={2021},
}

@article{xie2024adan,
  title={{Adan: Adaptive nesterov momentum algorithm for faster optimizing deep models}},
  author={Xie, Xingyu and Zhou, Pan and Li, Huan and Lin, Zhouchen and Yan, Shuicheng},
  journal={IEEE Transactions on Pattern Analysis and Machine Intelligence},
  volume={46},
  number={12},
  pages={9508--9520},
  year={2024},
}

@article{carroll2009optimizing,
  title={{Optimizing content-preserving projections for wide-angle images}},
  author={Carroll, Robert and Agrawala, Maneesh and Agarwala, Aseem},
  journal={ACM Transactions on Graphics},
  volume={28},
  number={3},
  pages={43},
  year={2009}
}

@ARTICLE{9503377,
  author={Bross, Benjamin and Wang, Ye-Kui and Ye, Yan and Liu, Shan and Chen, Jianle and Sullivan, Gary J. and Ohm, Jens-Rainer},
  journal={IEEE Transactions on Circuits and Systems for Video Technology}, 
  title={{Overview of the Versatile Video Coding (VVC) Standard and its Applications}}, 
  year={2021},
  volume={31},
  number={10},
  pages={3736-3764},
  doi={10.1109/TCSVT.2021.3101953}}

@article{matoba2019vr8k,
  title={{8K VR Video Live Streaming and Viewing System for the 5G Era}},
  author={Matoba, Naoto},
  journal={NTT DOCOMO Technical Journal},
  volume={20},
  number={4},
  pages={43--50},
  year={2019},
  publisher={NTT DOCOMO, INC.},
}

@article{eltobgy2020mobile,
  title={{Mobile streaming of live 360-degree videos}},
  author={Eltobgy, Omar and Arafa, Omar and Hefeeda, Mohamed},
  journal={IEEE Transactions on Multimedia},
  volume={22},
  number={12},
  pages={3139--3152},
  year={2020},
  publisher={IEEE}
}

@article{ling2025multi,
  title={{A Multi-Grid Implicit Neural Representation for Multi-View Videos}},
  author={Ling, Qingyue and Cheng, Zhengxue and Feng, Donghui and Wang, Shen and Zhu, Chen and Lu, Guo and Sun, Heming and Katto, Jiro and Song, Li},
  journal={arXiv preprint arXiv:2509.16706},
  year={2025}
}

@inproceedings{jia2025towards,
  title={{Towards practical real-time neural video compression}},
  author={Jia, Zhaoyang and Li, Bin and Li, Jiahao and Xie, Wenxuan and Qi, Linfeng and Li, Houqiang and Lu, Yan},
  booktitle={Proceedings of the Computer Vision and Pattern Recognition Conference},
  pages={12543--12552},
  year={2025}
}

@inproceedings{zhang2025flavc,
  title={{FLAVC: Learned Video Compression with Feature Level Attention}},
  author={Zhang, Chun and Sun, Heming and Katto, Jiro},
  booktitle={{Proceedings of the Computer Vision and Pattern Recognition Conference}},
  pages={28019--28028},
  year={2025}
}

@inproceedings{regensky2025beyond,
  title={{Beyond Perspective: Neural 360-Degree Video Compression}},
  author={Regensky, Andy and Windsheimer, Marc and Brand, Fabian and Kaup, Andre},
  booktitle={Proceedings of the IEEE/CVF International Conference on Computer Vision},
  pages={16143--16153},
  year={2025}
}

@article{arai2025neural,
  title={{Neural Compression of 360-Degree Equirectangular Videos using Quality Parameter Adaptation}},
  author={Arai, Daichi and Kondo, Yuichi and Unno, Kyohei and Sugito, Yasuko and Kusakabe, Yuichi},
  journal={arXiv preprint arXiv:2512.20093},
  year={2025}
}

@article{guo2025generative,
  title={{Generative Latent Video Compression}},
  author={Guo, Zongyu and Jia, Zhaoyang and Li, Jiahao and Zhang, Xiaoyi and Li, Bin and Lu, Yan},
  journal={arXiv preprint arXiv:2510.09987},
  year={2025}
}

\end{document}